\documentclass[11pt,letterpaper,twocolumn]{article}

\usepackage{times}
\usepackage[utf8x]{inputenc}
\usepackage[T1]{fontenc}
\usepackage{listings}

\usepackage[letterpaper,top=3cm,bottom=3cm,left=2cm,right=2cm,marginparwidth=1.75cm]{geometry}
\usepackage{amsmath}
\usepackage{graphicx}
\usepackage[colorinlistoftodos]{todonotes}
\usepackage[hyphens]{url}

\title{Simulating Coverage Path Planning with Roomba}
\author{Robert Chuchro\\
{\tt\small chuchro3@stanford.edu}}

\date{}

\begin{document}
\maketitle

\begin{abstract}
Coverage Path Planning involves visiting every unoccupied state in an environment with obstacles. In this paper, we explore this problem in environments which are initially unknown to the agent, for purposes of simulating the task of a vacuum cleaning robot. A survey of prior work reveals sparse effort in applying learning to solve this problem. In this paper, we explore modeling a Cover Path Planning problem using Deep Reinforcement Learning, and compare it with the performance of the built-in algorithm of the Roomba, a popular vacuum cleaning robot.
\end{abstract}

\section{Introduction}

	A Roomba is a robot vacuum cleaner built with sensors to automatically vacuum a living space while avoiding obstacles and stair ledges. The Roomba contains several sensors which allow it to receive information from the outside world. In this paper, the relevant sensors are the bumper sensors in front of the Roomba. This sensor will be triggered when the Roomba encounters an obstacle, such as a wall or piece of furniture. The Roomba can move by turning itself or driving forward. 

\section{Problem Statement}

The environment consists of a bounded two-dimensional plane which may contain obstacles. Any location which is not an obstacle is reachable from any other location in the environment. The goal is for the Roomba to traverse the entire open space in under an hour, avoiding collisions with obstacles.

Initially, the environment is unknown. The agent will need to adopt a policy to explore the space around it.

\subsection{Representation}
\begin{figure}
\centering
\includegraphics[width=0.3\textwidth]{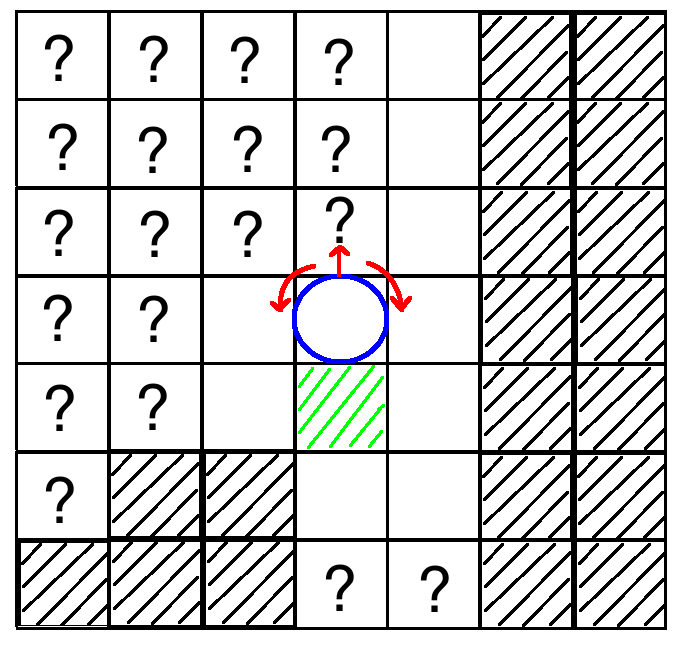}
\caption{\label{fig:gridworld}Grid world representation of Roomba's environment. The green cell is the base / start location.}
\end{figure}

To simplify the problem, we will discretize the environment as a grid world, with each grid cell being a square with length equal to the diameter of the Roomba, as shown in Figure ~\eqref{fig:gridworld}.

The agent will initialize its world as only having knowledge of the starting cell, with the remaining neighboring states as hidden states, represented with '?'. The agent has three actions available to it: drive forward, turn left 90 degrees, and turn right 90 degrees. All three actions are assigned to take one second, which is represented as one time step. 

When the agent attempts to move onto a hidden state, it will observe either an empty spaces or an obstacle, based on feedback from its bumper sensor. Here, the goal is to explore the hidden cells, ensuring that every empty space was visited.

\subsection{Success Measurement}
Success of a particular strategy can be measured by time taken to fully explore a given environment. In the context of simulating, this can be measured by the number of actions taken before visiting all empty cells.

In order to avoid modeling a desired or complex behavior through the reward function, the reward function remains kept simple:

\begin{equation}
\begin{split}
R(s, a, s')=\begin{cases}
1 & s'\text{ is an unvisited cell},\\
-1 & s'\text{ is an obstacle},\\
0 &\text{otherwise}
\end{cases}
\end{split}
\label{eq:long}
\end{equation}

The environment also has a natural decay, since we are limited to one hour (1800 time steps).

One of the goals is to come up with a strategy that can outperform Roomba's baseline algorithm, which will be outlined in a subsequent section.

Performances amongst the agents can be compared using the following metrics: percentage of open space explored, number of collisions with obstacles, and number of time steps to completion.

\section{Related Work}

\subsection{Path Planning using Discretized States}

The coverage algorithm described in ~\cite{Zelinsky93} involves discretizing a state space with obstacles, and planning safe paths betweens states. Based on simulation results, it was shown that using a 4-connected grid representation of the environment created paths that were no more than 2\% less efficient than if it were 8-connected. This means that restricting the turn angles of the agent to 90 degrees does not have a drastically negative effect on simulation results, in exchange for reducing the size of the state and action space.

\subsection{Boustrophedon Cellular Decomposition}
Boustrophedon or S-shaped pathing, follows the same pathing that farmers use on fields of crops, or lawn mowers on a field of grass. The algorithm involves finding a perimeter, then traversing back and forth along the length of the perimeter across the entire space ~\cite{hasan14}.

This pathing concept was extended in ~\cite{Choset1998} by attempting to divide the region into segments, each of which can be completed by executing Boustrophedon pathing.

\subsection{Combination of Basic Policies}

A combination of policies was shown to outperform any individual strategy in~\cite{hasan14}. The basic pathing strategies were:

\begin{itemize}
\setlength\itemsep{.2mm}
\item[-] Outward Spiral
\item[-] Wall Walking
\item[-] Random Walk
\item[-] Boustrophedon Pathing
\end{itemize}

It was not stated how the combination policy was determined; however, a possible approach would be to learn a scoring function to evaluate which strategy to use based on the current set of observations.

\subsection{Deep Reinforcement Learning}

One of the main challenges of this task is for the agent to learn a policy that generalizes well so that its policy is not specific to a particular environment.
A classic example of generalized reinforcement learning is presented in ~\cite{DBLP:journals/corr/MnihKSGAWR13}. Minh, et al. introduced a Deep Q-Network (DQN), a neural network used for Q-Learning. They used a convolutional neural network to train seven Atari 2600 games. They were able to achieve super-human results in six of the seven games it was tested on, with no adjustment of the architecture or hyperparameters between different games.

We will apply Deep Q-Learning to the simulated environment to produce an agent that can explore well in unseen environments.

\section{Methods}

\subsection{Baseline Roomba Algorithm}

The basic Roomba strategy consists of exploring outward in a spiral fashion until the nearest obstacle is reached. The Roomba then alternates between moving along a discovered obstacle edge and turning then driving forward until colliding with another obstacle~\cite{layton05}. The Roomba repeats this behavior for an hour, then attempts to return to its base.

In the context of our grid world representation, the turn angles will need to be restricted to cardinal directions. After colliding with an obstacle, the agent will then do random walks for the duration of the episode -- driving forward until an obstacle is encountered, then turning. The algorithm is outlined below:

\begin{lstlisting}[frame=single, language=Octave, numbers=right, numbersep=-5pt]
function baselineRoomba()
  initialize s, agent
  while not done
    if agent.usingSpiral
      a = spiral(agent, s)
    else
      a = randomWalk(agent, s)      
    s = observeNextState(s, a)                
\end{lstlisting}

\begin{figure}
\centering
\includegraphics[width=0.25\textwidth]{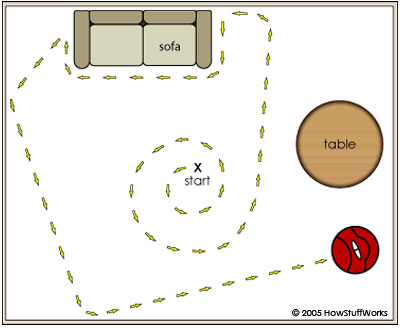}
\caption{\label{fig:baselinealgo} Depiction of Roomba's basic exploration algorithm \cite{layton05}.}
\end{figure}

\subsection{Deep Q-Learning}

Pseudo code for Q-Learning algorithm is as follows:
\begin{lstlisting}[frame=single, mathescape=true, language=Octave, numbers=right, numbersep=-5pt]
initialize Q[num_states,num_actions]
observe initial state s 
while condition
  select and carry out an action a 
  observe reward r and new state s'
  $Q[s,a]\ += \alpha(r + \gamma max Q[s',a'] - Q[s,a])$
  s=s'
\end{lstlisting}

$\alpha$ is the learning rate that controls how much is difference between previous Q-value and newly proposed Q-value. When $\alpha$ is equal to 1 then update equation is similar to Bellman equation.

For Deep Q network given a transition $<s,a,r,s'>$, the Q-table update rule can be re written as: 
\begin{itemize}
\setlength\itemsep{1mm}
\item[] Do a feedforward pass for the current state s to get predicted Q-values for all actions.
\item[] Do a feedforward pass for the next state s’ and calculate maximum overall network outputs $maxQ(s’, a’)$.
\item[] Set Q-value target for action to $r + \gamma * maxQ(s’, a’)$. For all other actions make output equal to 0.
\item[] Update the weights using back propagation (gradient descent).
\item[] Optimize with L2 loss: 
$Loss = (r + maxQ(s',a') - Q(s,a))^2 $
\end{itemize}

Our implementation of Deep Q-Learning is implemented using Tensor Flow ~\cite{tensorflow2015-whitepaper}. The programming language used was Julia-0.6, and we utilized a Tensor Flow wrapper library ~\cite{tensorflow-julia}.
We use Tensor Flow’s Adam optimization algorithm with an initial learning rate of 2e-4.

\subsection{Network Architecture}

It is typical that a neural network will consist of several convolutional layers, followed by fully connected layers, such as in ~\cite{DBLP:journals/corr/MnihKSGAWR13}. However, in image recognition problems, the input data is typically very large and the convolutional layers are seen as feature identifiers, whereas the fully connected layers take in the features from previous layers and use their weights to identify correlations or non-linear combinations between the input features ~\cite{Deshpande}.

The input to our network which represents the state consists of two components: the first is the number of time steps lapsed, and the second is a $9 \times 9$ array consisting of the partially observed surrounding environment. That is, the center is marked by the current location, while the surrounding cells consist of the following values:

\begin{itemize}
\setlength\itemsep{.2mm}
\item[-1]  : Unobserved cell
\item[0]   : Observed empty cell
\item[1]   : Observed obstacle
\item[2]   : Base/Starting cell
\end{itemize}

By restricting our input to a fixed subset of the actual observed environment, we are not allowing the model to focus on specific features of an environment, but instead on learning to make decisions based on potentially reoccurring subsets of environments.

Since the input to our network is already fairly structured, we omit the convolutional layers and use only two fully connected layers, the first with width 64, and the second with a width equal to the number of actions, which in this simulation is 3. These outputs represent our Q-values. The fully connected layers are separated by a ReLU non-linearity activation. ~\cite{pmlr-v15-glorot11a}.

\subsection{Generating Random Environments}

\begin{figure}
\centering
\includegraphics[width=0.35\textwidth]{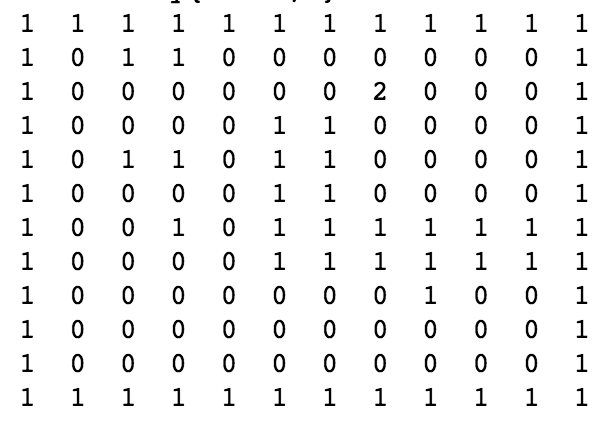}
\caption{\label{fig:sample-env} Array representation of sample environment generation.}
\end{figure}

Each simulation episode generated a unique environment, so that agents could not simply learn how to navigate a specific layout. The environment generation algorithm first generated a 2D room layout with dimensions ranging between 10 and 20. Then, out of 17 'furniture' pieces, up to 6 were randomly placed inside the layout with random orientation. A sample layout is depicted in figure ~\eqref{fig:sample-env}.

\section{Experiments}

\subsection{Sinusoidal Exploration Decay}

Typically, reinforcement learning algorithms begin with a high exploration rate, and decrease it as training progresses. This models the behavior of initially exploring, followed by exploiting as the model learns the environment. The decay method is traditionally a linear or exponential decay. In our early trials, we found that after a certain number of iterations, training would progress would flatten, but then continue to improve when a new epoch was started. When a new epoch begins, the most significant difference in our implementation is the reset of the exploration hyperparameter, $\epsilon$. 
We have introduced a new $\epsilon$ decaying function: one which exponentially decays over episodes in a sinusoidal fashion.

\begin{equation}
\begin{split}
\epsilon = \epsilon_0 \cdot \epsilon_d^x \cdot \frac{1}{2}(1 + cos(\frac{2 \pi x n}{X}))
\end{split}
\label{eq:long}
\end{equation}

\begin{itemize}
\setlength\itemsep{.2mm}
\item[-] $\epsilon_0$ is initial epsilon
\item[-] $\epsilon_d$ is decay rate
\item[-] n is number of mini epochs
\item[-] X is number of training episodes
\item[-] x is current training episode number
\end{itemize}

\begin{figure}[h]
\begin{center}
\includegraphics[width=.4\textwidth]{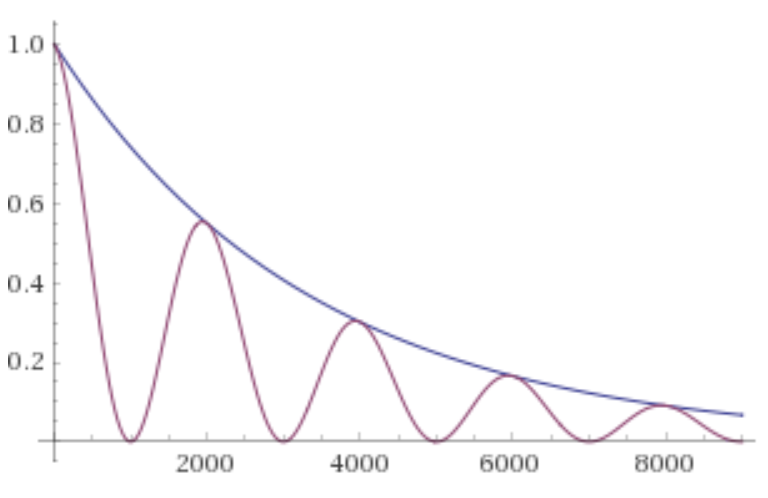}
\caption[]{Plot of equation ~\eqref{eq:long} (purple) alongside with $\epsilon^x$ (blue). The model can be saved at the troughs of the waves, serving as an epoch. Parameters used: X=10000, n=5, $\epsilon_0$=1, $\epsilon_d$=.9997}
\end{center}
\end{figure}

The main motivation behind this function was for our model to be able to escape local optima while training with an already reduced $\epsilon$. Since it is difficult to model an ideal exploration rate for a particular environment, a decaying sinusoidal function makes less assumptions about the expected decay of $\epsilon$, hopefully finding the correct decay rate periodically during the entire epoch. An analogy can be made about the suggested strategy. It focuses on the idea how one should use car brakes on a slippery road (without ABS): it is difficult for a driver to apply the brakes an optimal amount while maintaining rolling friction, but if the driver pumps the brakes repeatedly, they will achieve the correct amount of pressure periodically.

\subsection{Baseline vs DQN}

\begin{figure}
\centering
\includegraphics[width=0.4\textwidth]{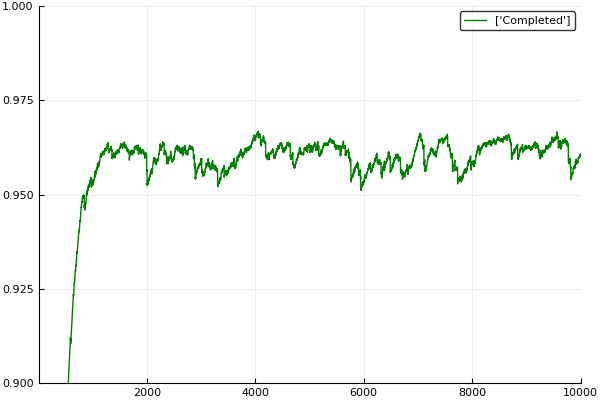}
\includegraphics[width=0.4\textwidth]{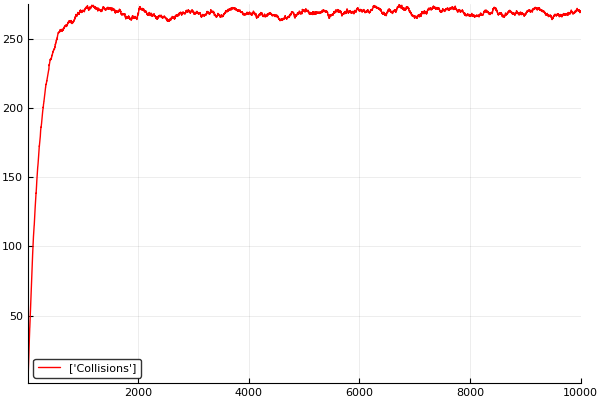}
\caption{\label{fig:baseline-graphs} Running averages of percent coverage (green) and number of collisions (red) over 10000 episodes for the baseline algorithm}
\end{figure}

Despite the simplicity and randomness of Roomba's basline algorithm, a simple combination of spiraling and straight walks with random turns demonstrated to be fairly thorough in coverage; however, due to the randomness, it frequently collided with obstacles. Since no learning occurred, results were consistent across the 10000 iterations ~\eqref{fig:baseline-graphs}.

\begin{figure}
\centering
\includegraphics[width=0.4\textwidth]{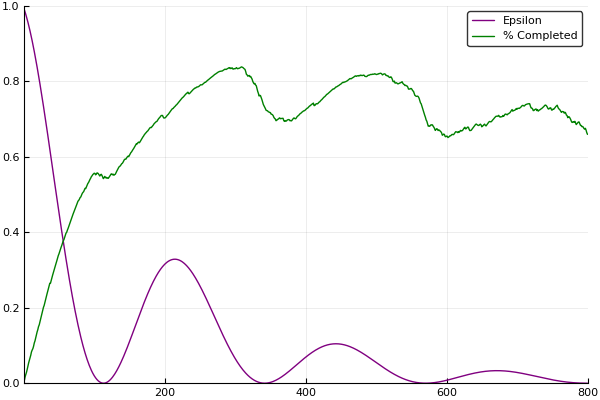}
\includegraphics[width=0.4\textwidth]{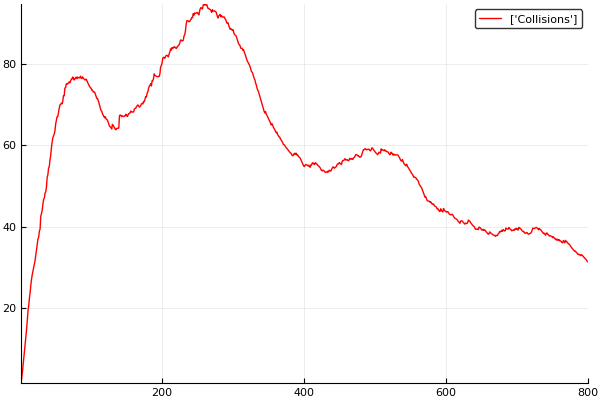}
\caption{\label{fig:dqn-graphs} Running averages of percent coverage (green) and number of collisions (red) over 800 episodes for the DQN}
\end{figure}

Preliminary results for only 800 episodes of training using Deep Q-Learning indicate a strong ability of learning to avoid obstacles, however, there appears to be a policy trade off between minimizing collisions and coverage percentage ~\eqref{fig:dqn-graphs}.

Table ~\eqref{fig:perf-table} compares coverage percentage and obstacle avoidance across several preliminary epochs of the DQN as well as the Roomba baseline algorithm. At one point, the DQN is able to get over 80\% coverage, however, this decreases as the agent learns to avoid obstacles. It is also worth noting that coverage is better when $\epsilon$ is slightly greater than 0, indicating that some randomness may assist with exploring an unknown environment.

\begin{figure}
\begin{center}
{\scriptsize
\begin{tabular}{ |c||c|c| }
 \hline
 \multicolumn{3}{|c|}{Performance Comparison} \\
 \hline
 Agent & \% Coverage & \# of Collisions\\
 \hline
 \hline
 Baseline Roomba & 96 & 268\\
 \hline
 DQN-e1 & 55 & 64\\
 \hline
 DQN-e2 & 72 & 63\\
 \hline 
 DQN-e3 & 74 & 49\\
 \hline
 DQN-e4 & 66 & 32\\
 \hline
\end{tabular}
}
\end{center}
\caption{\label{fig:perf-table} An apparent trade off between coverage and collisions.}
\end{figure}

\section{Conclusion}

Preliminary results show that Deep Q-Learning is effective at learning to avoid obstacles, and that the graphs had not yet leveled off after 800 iterations of training. This warrants for further exploration of this model, and more training. One Limitation of this study is the insufficient number of training iterations for the DQN, which should be well into the thousands, instead of 800. As prior studies suggest, it is likely difficult to properly model this problem with a pure learning algorithm, and a hybrid approach of applying some variation of a search algorithm with learning will likely prove to be more effective.

\bibliographystyle{plain}
\bibliography{sample}

\end{document}